\documentclass[10pt,conference]{IEEEtran}

\usepackage{hyperref}
\usepackage{multirow}
\usepackage{xcolor}
\usepackage{booktabs}
\usepackage{cite}
\ifCLASSINFOpdf
   \usepackage[pdftex]{graphicx}
   \graphicspath{{figs/}}
   \DeclareGraphicsExtensions{.pdf,.jpeg,.png}
\else
   \usepackage[dvips]{graphicx}
   \graphicspath{{../figs/}}
   \DeclareGraphicsExtensions{.eps}
\fi
\usepackage{lipsum}

\newcommand\blfootnote[1]{%
  \begingroup
  \renewcommand\thefootnote{}\footnote{#1}%
  \addtocounter{footnote}{-1}%
  \endgroup
}

\usepackage[cmex10]{amsmath}
\usepackage{amsthm}

\usepackage{array}
\ifCLASSOPTIONcompsoc
  \usepackage[caption=false,font=normalsize,labelfont=sf,textfont=sf]{subfig}
\else
  \usepackage[caption=false,font=footnotesize]{subfig}
\fi
\usepackage{url}
\newif\iffinal
\finalfalse
\finaltrue
\iffinal
\else
\fi

\begin{document}
\title{Neuroevolution-based Classifiers for Deforestation Detection in Tropical Forests}


\iffinal

\author{\IEEEauthorblockN{Guilherme A. Pimenta*}
\IEEEauthorblockA{Institute of Science and Technology \\ Universidade Federal de S\~{a}o Paulo \\  S\~{a}o Jos\'{e} dos Campos, Brazil \\ gpimenta@unifesp.br }
\and
\IEEEauthorblockN{Fernanda B. J. R. Dallaqua}
\IEEEauthorblockA{Visiona Tecnologia Espacial S.A.\\  S\~{a}o Jos\'{e} dos Campos, Brazil \\ fernanda.dallaqua@visionaespacial.com.br}
\and
\IEEEauthorblockN{\'{A}lvaro Fazenda and Fabio A. Faria*
\IEEEauthorblockA{Institute of Science and Technology \\ Universidade Federal de S\~{a}o Paulo \\  S\~{a}o Jos\'{e} dos Campos, Brazil \\ \{alvaro.fazenda,ffaria\}@unifesp.br}
}}

\maketitle

\begin{abstract}
Tropical forests represent the home of many species on the planet for flora and fauna, retaining billions of tons of carbon footprint, promoting clouds and rain formation, implying a crucial role in the global ecosystem, besides representing the home to countless indigenous peoples. Unfortunately, millions of hectares of tropical forests are lost every year due to deforestation or degradation. To mitigate that fact, monitoring and deforestation detection programs are in use, in addition to public policies for the prevention and punishment of criminals. These monitoring/detection programs generally use remote sensing images, image processing techniques, machine learning methods, and expert photointerpretation to analyze, identify and quantify possible changes in forest cover. Several projects have proposed different computational approaches, tools, and  models to efficiently identify recent deforestation areas, improving deforestation monitoring programs in tropical forests. In this sense, this paper proposes the use of pattern classifiers based on neuroevolution technique (NEAT) in tropical forest deforestation detection tasks. Furthermore, a novel framework called e-NEAT has been created and achieved classification results above $90\%$ for balanced accuracy measure in the target application using an extremely reduced and limited training set for learning the classification models. These results represent a relative gain of $6.2\%$ over the best baseline ensemble method compared in this paper~\blfootnote{\textbf{*The G. A. Pimenta and F. A. Faria contributed equally to this work (co-first authorship).}}.
\end{abstract}

\IEEEpeerreviewmaketitle

\section{Introduction}
\label{s.intro}
Tropical forests are located between the tropics of Capricorn and Cancer, close to the equator. They have hot and humid weather, great biodiversity (more than half of the species of the planet) besides regulating the climate and rainfall through their plants and trees that also absorb large amounts of carbon dioxide~\cite{AFW14,martin2015edge}. 

Unfortunately, tropical forests suffer from indiscriminate and progressive deforestation and degradation due to livestock, agriculture, wood extraction, or urban area expansion~\cite{AFW14}. According to INPE (Brazilian's National Institute for Space Research), the deforestation in the Brazilian Legal Amazon (BLA) between the period of August 1, 2019, and July 31, 2020, was $10,851 km^2$, an increase of $7.13\%$ considering the previous period~\cite{prodes2021}. This increasing deforestation can bring irreversible and catastrophic consequences, such as loss of biodiversity, climate change, desertification, water scarcity, increase in diseases, and even the emergence of pandemics~\cite{martin2015edge}.

Monitoring programs by government agencies and non-profit institutions were created to help the conservation of tropical forests. These programs use remote sensing imagery, image processing, machine learning methods, and experts' photo-interpretation to analyze, identify and quantify changes in the environment~\cite{AFW14}. Some monitoring programs are PRODES, DETER, SipamSAR, SAD, and GLAD. 

PRODES (Amazon Deforestation Monitoring Project), a program created in 1988 by the INPE, uses experts' photo-interpretation of satellites imagery to map and calculate the annual deforestation rate of BLA. Most of the analyzed images are from Landsat satellites~\cite{guide2017landsat}, providing $30$ meters of spatial resolution and 16 days of revisit time. PRODES methodology can be divided into three steps: (1) image selection and enhancing of the scenes with little cloud cover acquired near August, 1;  (2) mapping of deforestation polygons by the experts' photo-interpretation; and (3) calculation of the annual deforestation rate~\cite{metodprodes2019}. The classified mosaic and deforestation rates come from the TerraBrasilis platform~\cite{fg2019terrabrasilis}. PRODES data don't infer on regenerated areas, therefore deforested segments are included in an exclusion mask used in subsequent years. This mask also takes into account different vegetation and hydrography areas.

DETER program, also held by INPE, was created in $2004$ to stop or mitigate the beginning or extension of a deforestation process.
Images with little cloud cover are daily collected with $64$ meters of spatial resolution from sensor WFI/CBERS-4. Soil, shadow, and vegetation fractions are estimated through a Linear Spectral Mixing Model, considering an image composed of red, near-infrared, and green bands. Soil and shade fractions are photo-interpreted and analyzed by experts users implying possible alerts sent to the authorities~\cite{metodprodes2019}. To complement DETER alerts, the Ministry of Brazil created the program SipamSAR~\cite{teixeira2015uso,sipamsar2007} which uses data from Synthetic Aperture Radar, enabling the detection of deforestation in areas and periods with high cloud cover.  

SAD and GLAD programs are maintained by the non-profit institutions Imazon and Global Forest Watch, respectively. The first analyzes the pixels from optical and radar images of Landsat and Sentinel satellites according to a threshold of a spectral index called Normalized Difference Fraction Index (NDFI)~\cite{SOUZA2005329,IMAZON2009}. The estimates are available every month on the program's website. GLAD program uses Landsat images and bagged decision trees to calculate the pixel's median likelihood of forest cover loss which is thresholded at higher than $50$\%, generating an alert. This program also monitors other tropical forests such as Congo Basin and Southeast Asia~\cite{GLAD2016}.

Recently, the ForestEyes  project~\cite{foresteyes2019,dallaqua2021foresteyes,Dallaqua_GRSL2022} was proposed as an alternative solution to support monitoring program of detecting deforestation in tropical forests. It uses classifiers based on support vector machines (SVM), however it doesn't use PRODES labeling as groundtruth for the training set. Instead, it applies Citizen Science~\cite{GREY2009,SILVERTOWN2009} concept, which uses volunteers to classify remote sensing images from Landsat-8 into Forest or Non-forest and the majority vote of these volunteer classifications to define the final label of the samples. The main goal of ForestEyes is to train classifiers with small but efficient datasets to detect deforestation in tropical forests, complementing data for official monitoring programs or generating data where this kind of program does not exist. 

In literature, other initiatives have been proposed to improve deforestation detection tasks which could make the monitoring more agile, helping the official programs and the experts. Some of these initiatives rely on traditional machine learning methods~\cite{zhu2014continuous,schneider2012monitoring,Dallaqua_GRSL2022} and more recently, deep learning architectures~\cite{de2020change,rs12060910,rs13245084}. 

Several deep learning techniques were compared in~\cite{de2020change,rs12060910,rs13245084} for detecting deforestation in BLA. Images from Landsat-8~\cite{de2020change,rs12060910,rs13245084} and Sentinel-2~\cite{rs13245084} were used as input of the models and PRODES data were used as groundtruth. The results look promising however~\cite{de2020change} found that the performance is inferior with small datasets which could complicate the process of building a model for broader use. Another challenge is the detection of recent deforestation in the Amazon, as remaining vegetation commonly found in its deforestation process hinders detection~\cite{rs12060910}.

Aiming to improve the classification results in deforestation detection tasks using a reduced training set, this paper proposes the use of ANN classifiers based on the neuroevolution technique as well as a framework for combining those ANN classifiers, which might provide complementary information to the final ensemble of classifiers.

%In sense to improve the classification results in deforestation detection tasks using reduced training set, this paper proposes a framework for combining ANN classifiers based on neuroevolution technique. Furthermore, four different aggregation strategies for combining classifiers are developed aiming to select the more suitable ANNs, which they might provide complementary information to the final ensemble of classifiers.

\section{Ensemble of Classifiers based on Neuroevolution Technique}
\label{s.proposal}

The ensemble of classifiers consists of combining base classifiers focusing on solving a target problem. They use several combined classifiers to create a unique and stronger model, allowing them to accomplish more effective learning~\cite{Hansen:90}. The generalization ability of an ensemble method is usually higher than the base classifiers, due to the increase in the diversity of features extracted and decisions made~\cite{Schapire:90}. An ensemble of classifiers is composed of several base classifiers such as decision trees (DT), support vector machines (SVM), and artificial neural networks (ANNs). 

A successful technique to create ANNs incrementally through evolutionary processes is called NEAT (NeuroEvolution of Augmenting Topologies). NEAT operates by modifying both connection weights and network structure, to design low-complexity networks. Such an algorithm proved to be successful mainly when applied to tasks involving reinforcement learning~\cite{neatArticle}.

In the process of evolving ANNs through the neuroevolution technique, different iteractions are perform during the evolutionary steps, resulting in networks with distinct topologies. Since diversity between base classifiers is deemed to be a key factor in the ensemble of classifiers, this paper proposes a framework for combining NEAT-based ANNs called e-NEAT (Ensemble based on NEAT) to achieve better results when compared with other single or multiple classification models.

%\faf{qual a motivacao para utilizar NEAT como base classifier no ensemble e nao outros modelos de aprendizagem (e.g., SVM)?}
e-NEAT consists of an ensemble method composed of a set of ANNs created through the neuroevolution technique implemented by the NEAT technique. 
Each ANN belonging to e-NEAT is created by targeting topologies that differ from each other so that the vote of the set when combined with some aggregation strategy can present better classification results than the vote of a single ANN. To address the diversity among the ANNs, e-NEAT focuses on two steps in the construction of the ensemble model: (1) the creation of the ANNs; and  (2) the vote aggregation strategy. Figure~\ref{Fig:eNeatDiagram} illustrates the e-NEAT method proposed in this work.

\subsection{Steps of the e-NEAT Approach}

The steps for building the final e-NEAT model are presented in Figure~\ref{Fig:eNeatDiagram} through the numbers incorporated into the connections between the blocks, in total there are five steps detailed in the following sections.

\subsubsection{\textbf{Splitting the Dataset}} 
the first step of the method consists of creating the two subsets (training and test). In this work, the deforestation dataset has already been defined in Section~\ref{sec:dataset}. However, in the other dataset, this step occurs in stratified sampling, meaning that the proportion between the classes in each subset is maintained. The number of samples in the training and test subsets is defined by the user in a crossvalidation protocol.%correspond to $85\%$ and $15\%$ of the original training set.
\begin{figure}[!ht]
\centering
\includegraphics[width=0.438\textwidth]{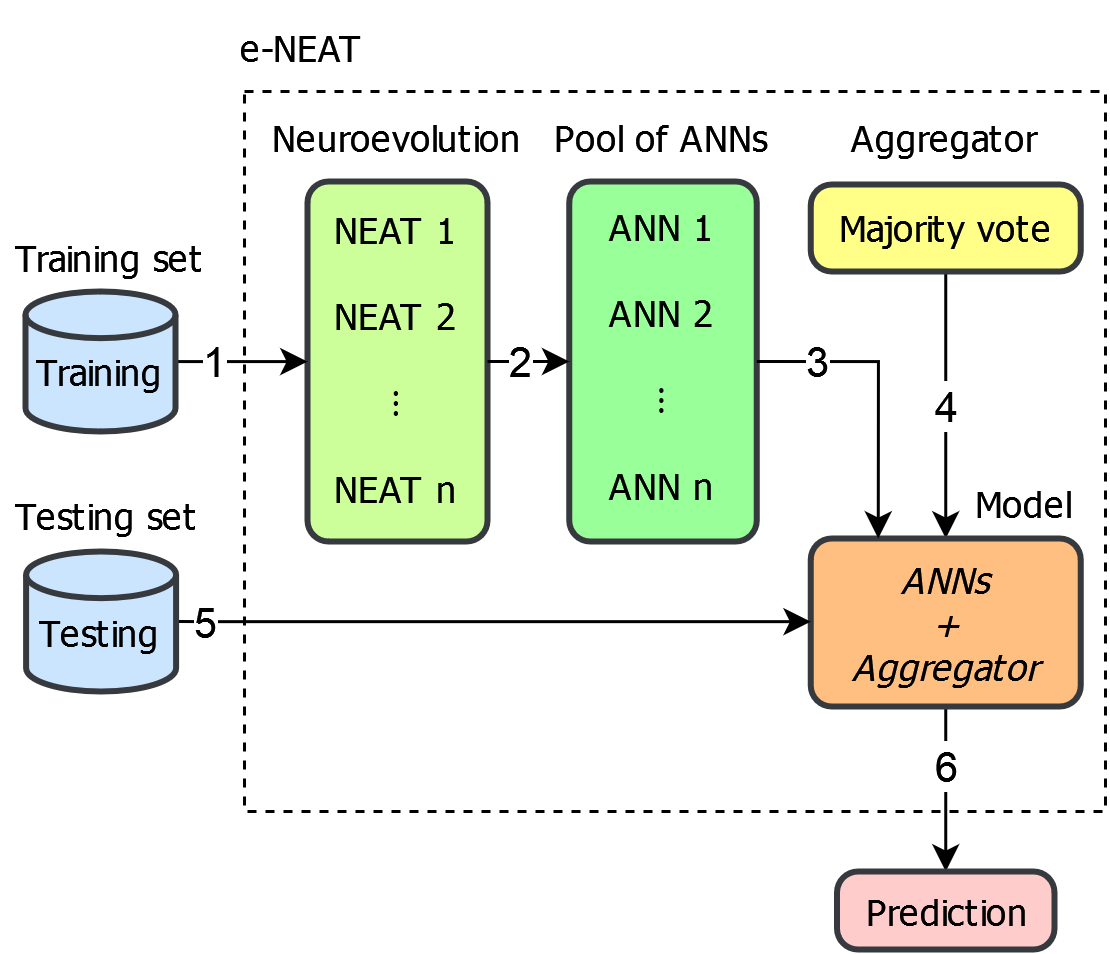}
\caption{Diagram of the e-NEAT method proposed in this work.}
\label{Fig:eNeatDiagram}
\end{figure}

\subsubsection{\textbf{Neuroevolution using NEAT Technique}}

The NEAT technique has been introduced by Kenneth O. Stanley and Risto Miikkulainen~\cite{neatArticle}. First, an initial population of ANNs is created. The second step consists in calculating a measure of structural similarity between the ANNs to group them into species that will compete primarily within their niches. Next, different genetic operations (e.g., selections, reproduction, and mutation) are applied to create new artificial neural networks (ANNs) that will compose a new population. This process is repeated until stop criteria (e.g., the fitness threshold or the number of generations) is reached. Finally, the best neural network is found and trained as the NEAT outcome.

%\begin{figure*}[!ht]
%\centering
%\includegraphics[width=0.8\textwidth]{figs/neat.png}
%%\caption{Evolving artificial neural network (ANN) using NEAT technique proposed in~\cite{neatArticle}}
%\label{Fig:neatDiagram}
%\end{figure*}

The objective of this step is to create neural networks with simplified topologies as well as they can achieve good classification results to the target task. Therefore, the method used by the NEAT technique already covers the evolution of ANNs gradually, in which the complexity of the final  neural network is proportional to the difficulty of the problem exposed to the NEAT technique.

To promote greater diversity among the neural networks present in the ensemble method (e-NEAT), some parameters of NEAT technique are changed during the creation of each ANN, such as population size and the set of available activation functions. Consequently, the neuroevolution process will tend to establish distinct paths during evolutionary steps, since the components available for the creation of the topology and the conditions are not the same.

During the creation of the final ensemble model (e-NEAT), \textit{n} different instances of NEAT technique are initiated, corresponding to the number of ANNs that will compose the final e-NEAT approach. These instances can occur either sequentially or in parallel, thus each artificial neural network is the product of an independent evolution process. At the end of this step, the model will have \textit{n} ANNs of different topologies, which they are combined by an aggregation strategy to provide the final classification of the ensemble method.

As fitness function (evaluation criteria), the balanced accuracy measure is adopted in order to guide the evolutionary process used in the NEAT technique. The main parameters assigned to this step consist of defining the fitness value and the maximum number of generations.

According to Kenneth O. Stanley and Risto Miikkulainen~\cite{neatArticle}, due to NEAT technique adds topological elements (nodes and connections) to the ANN throughout its training, it is less susceptible to get trapped in local optimum.

\subsubsection{\textbf{Creating the Pool of ANNs}} at the end of the neuroevolution process, the ANN with the best performance during training is extracted in the last generation. This procedure is performed for $\textit{n}$ different instances of NEAT technique, resulting in a pool of \textit{n} ANNs for being combined by an aggregation strategy (aggregator).

\subsubsection{\textbf{Classifier Aggregation Strategies (Aggregator)}} this step defines an aggregator that combines the ANNs created in the previous step, in order to achieve better classification results when the model is exposed to unseen samples. The choice of the aggregation strategy is related to its performance on the validation subset, using balanced accuracy as evaluation measure. In this work, the aggregation strategy adopted is the mode of different opinions of base classifiers. This strategy assigns the classification of the model to the class that received the highest number of votes (majority voting) from the $n$ ANNs. In case of tie, the strategy randomly grants the final vote within the set that presented the same number of votes. 

\subsubsection{\textbf{Building the Final Ensemble Model}}
Since the process of creating the pool of ANNs and the definition of the aggregation strategies during training have been performed, the final ensemble model is ready to predict the class of the unseen samples.

\section{Experimental Methodology}
\label{s.method}
This section describes the experimental methodology adopted in this paper.

\subsection{Dataset}
\label{sec:dataset}
%The dataset used in this work came from the ForestEyes project~\cite{foresteyes2019,dallaqua2021foresteyes,Dallaqua_GRSL2022}. 
The dataset consists of images from $5$ areas of the State of Rond\^{o}nia (Brazil)~\cite{Dallaqua_GRSL2022}. Figure~\ref{fig:studyarea} shows the study areas where one area was used as training set (in red) and the others were used as test set (in blue). As the areas belong to BLA, PRODES mosaic from the year of $2017$ was used to build the groundtruth images and the remote sensing images (Figure~\ref{fig:gt}(a)) were the same analyzed by INPE's experts which means Landsat-8 images with little cloud cover acquired near July 31, 2017.

%\begin{figure}[ht!]
%\begin{center}
%\includegraphics[width=0.45\textwidth]{figs/studyarea.pdf}
%\end{center}
%\caption{Five areas from State of Rond\^{o}nia (Brazil) used in this work.}
%\label{fig:studyarea}
%\end{figure}

\begin{figure}[ht!]
\begin{center}
\includegraphics[width=0.45\textwidth]{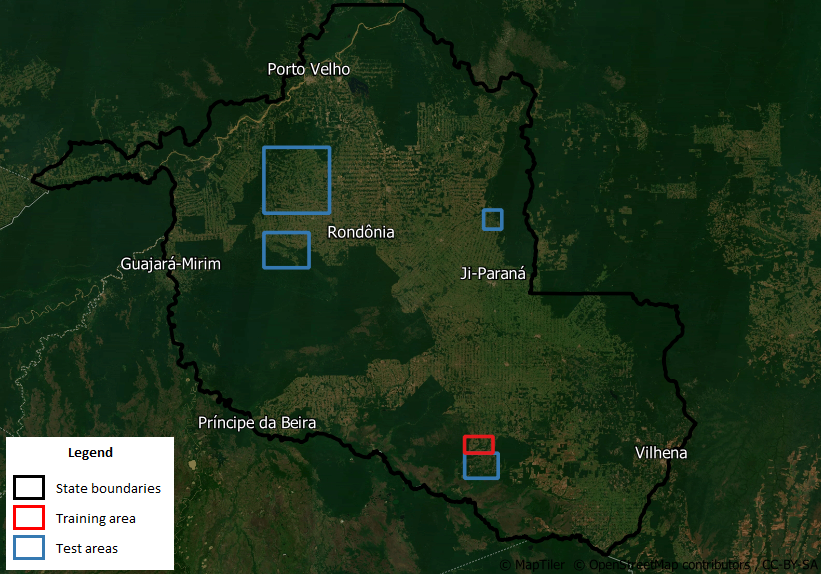}
\end{center}
\caption{The five areas from State of Rond\^{o}nia (Brazil) used in this work.}
\label{fig:studyarea}
\end{figure}

\begin{figure*}[!ht]
\begin{center}
\begin{tabular}{cc}
\includegraphics[width=0.5\textwidth]{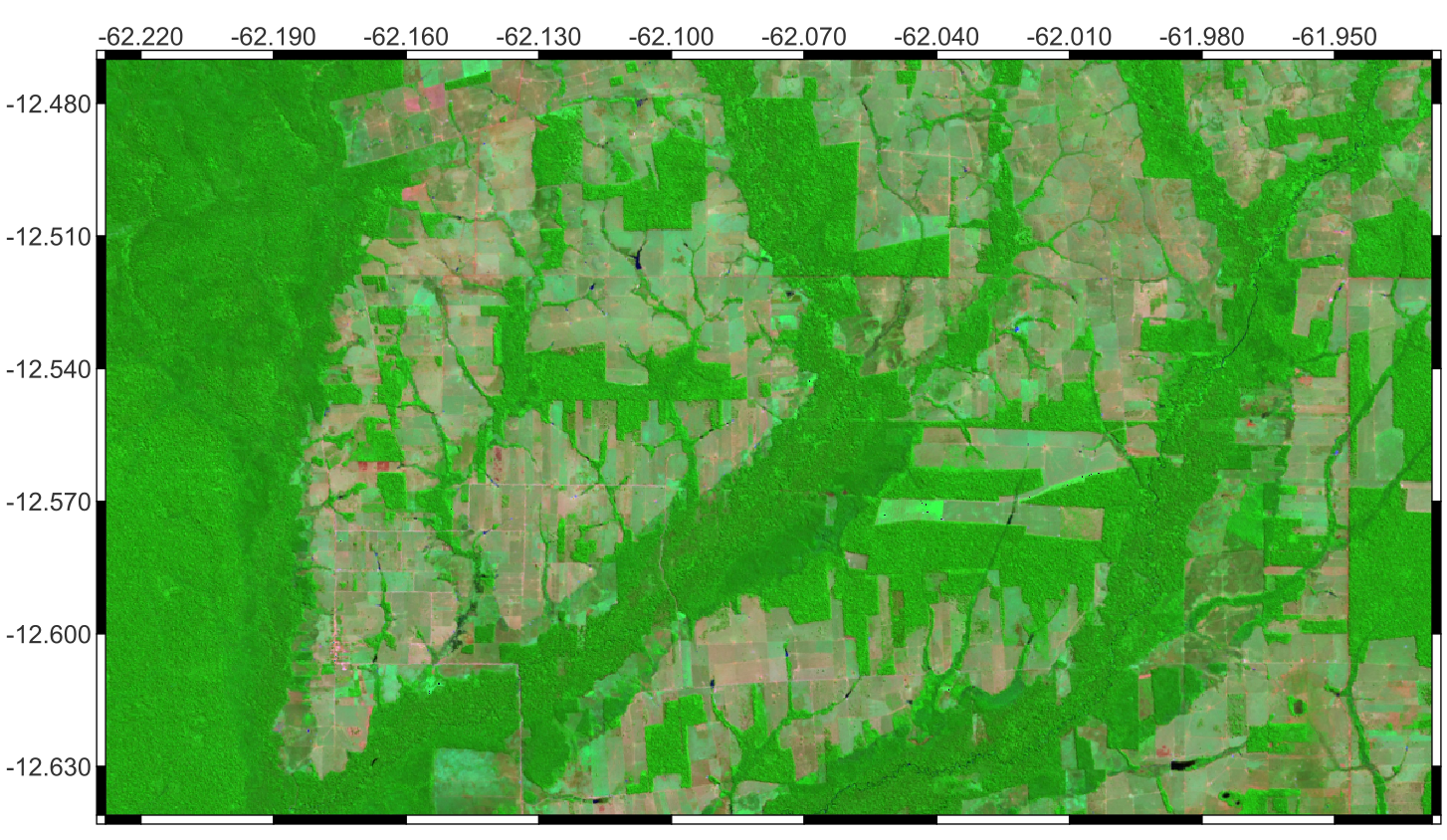} & \includegraphics[width=0.48\textwidth]{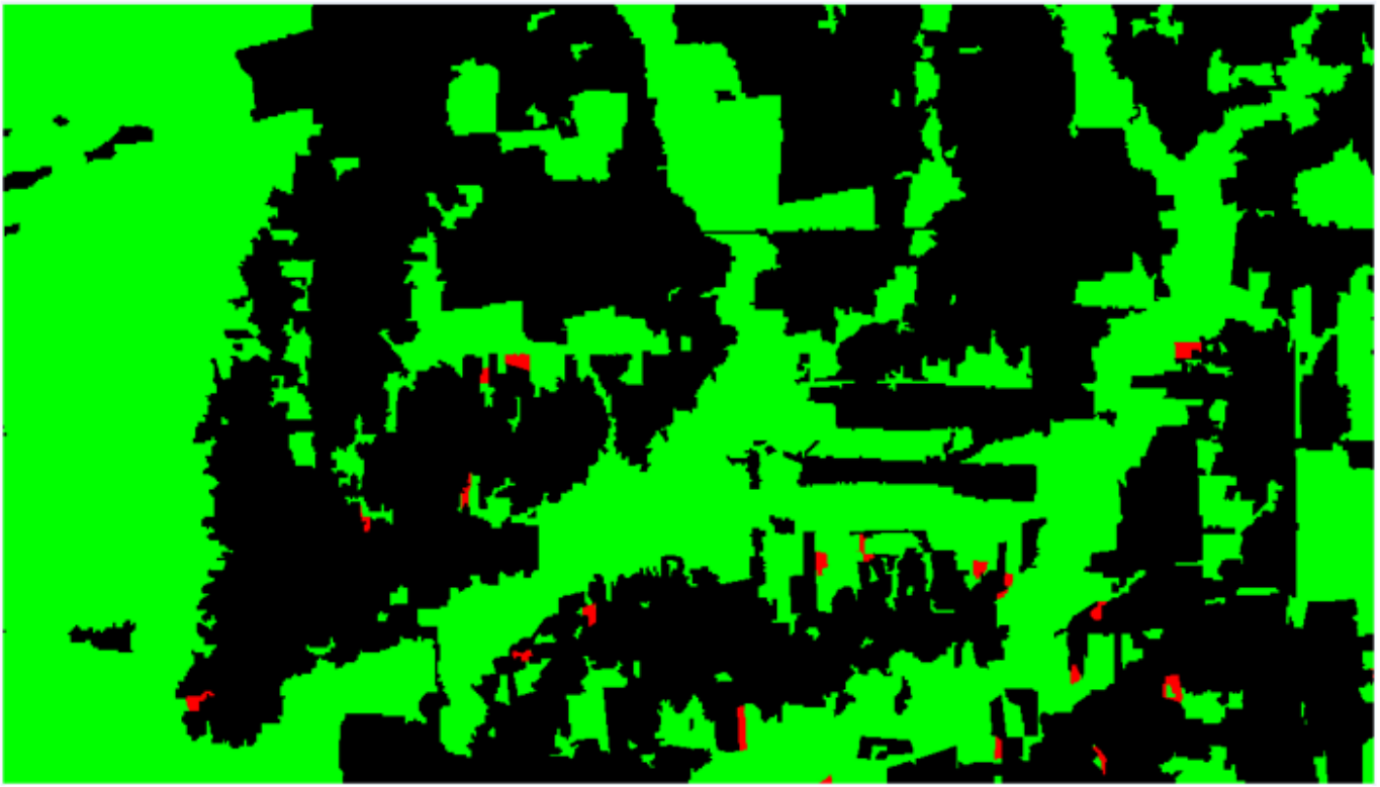} \\ 
(a)  & (b)  \\
\end{tabular}
\end{center}
\caption{The regions of interest (ROIs) used in this paper. In (a) is a false-composition of the remote sensing image and in (b) is a groundtruth image based on PRODES mosaic of $2017$, which red regions are Non-forest class, green regions are Forest class and black regions are regions no relevant for this paper.}
\label{fig:gt}
\end{figure*}

As this paper focuses on detecting recent deforestation area, a mask was created to exclude the pixels that received PRODES' classification as deforested before August 1, $2016$ (the black regions in Figure~\ref{fig:gt}(b)). The pixels outside the exclusion mask were segmented by MaskSLIC~\cite{irving2016maskslic} algorithm with the inputs being a 3-band image built by dimensionality reduction (Principal Component Analysis, PCA~\cite{PCA}) of $7$ Landsat-8 bands and the desired number of segments, which was defined as the number of pixels outside the mask divided by the minimum size area of deforestation tracked by PRODES ($6.25$ hectares, which corresponds to approximately $70$ pixels). 

In this paper, the segments are classified as Forest (F) or Non-Forest (NF). The label of each segment is defined by the majority class of its pixels. So, if a segment has $10$ Forest pixels and $5$ Non-Forest pixels, it will have Forest label. Table~\ref{tab:nsegments} presents the number of segments for each area. Both training and test areas are unbalanced since %the Amazon forest on those areas, for now, remains mostly preserved, and 
the exclusion mask covered all Non-forest areas before August 1, 2016.

For the training subset just a few segments were collected from the training area: $29$ segments that only present Non-Forest pixels, segments that are mainly composed by Forest pixels, and some segments that only present Forest pixels. Therefore, in this paper, a reduced training scenario has been adopted meaning that each learning technique has only $91$ segments to train a learning model and to classify $57,646$ segments on the test set. Table~\ref{tab:nsegments} presents the number of Non-Forest and Forest segments for each area of the dataset. %For the training subset just a few segments were collected from the training area: the $29$ Non-Forest segments, the Forest segments that have Non-Forest pixels and some Forest segments that only presents Forest pixels. Therefore, in this paper, a reduced training scenario has been adopted meaning that each learning technique has only $91$ segments to train a learning model and to classify $57646$ segments on the test set.

\begin{table}[ht!]
\caption{Dataset of image segments for recent deforestation detection proposed in~\cite{Dallaqua_GRSL2022}.}
\begin{tabular}{|c|ccccc|}
\hline
\multirow{2}{*}{\textbf{Classes}} & \multicolumn{5}{c|}{\textbf{Areas}}                                                                                                       \\ \cline{2-6} 
                         & \multicolumn{1}{c|}{\textbf{Training}} & \multicolumn{1}{c|}{\textbf{Test 1}} & \multicolumn{1}{c|}{\textbf{Test 2}} & \multicolumn{1}{c|}{\textbf{Test 3}} & \textbf{Test 4} \\ \hline
Forest (F)               & \multicolumn{1}{c|}{62}         & \multicolumn{1}{c|}{7,767}       & \multicolumn{1}{c|}{3,166}       & \multicolumn{1}{c|}{33,613}       &   8,556     \\ \hline
Non-Forest (NF)           & \multicolumn{1}{c|}{29}         & \multicolumn{1}{c|}{167}       & \multicolumn{1}{c|}{487}       & \multicolumn{1}{c|}{3,086}       &    804    \\ \hline
%\multirow{2}{*}{
\textbf{Total}%} %                   & %multicolumn{1}{c|}{91}         & \multicolumn{1}{c|}{7,934}       & \multicolumn{1}{c|}{3,653}       & \multicolumn{1}{c|}{36,699}       &  9,360      \\ \cline{2-6}
& \textbf{91} & \multicolumn{4}{|c|}{\textbf{57,646}} \\   \hline  
\end{tabular}
\label{tab:nsegments}
\end{table}

Furthermore, due to the reduced size of the training set, more complex description approaches in the literature (e.g., deep learning architectures) were not considered in this paper. Therefore, the well-known Haralick texture descriptor has been chosen as image descriptor to code the visual properties of the segments into feature vectors. Thirteen Haralick's coefficients were extracted for each Landsat-8 band. An exhaustive study has been conducted in~\cite{Dallaqua_GRSL2022} comparing $135$ band combinations to define which bands should to be adopted to represent those segments. The best results were obtained by the combination of Landsat-8 bands $b4$ (red) and $b6$ (shortwave infrared I).

%The $13$ Haralick's coeficients were extracted from Landsat-8 using $b4$ (red) and $b6$ (shortwave infrared I) bands, resulting in a feature vector with $26$ dimensions. An exhaustive study has been conducted in~\cite{Dallaqua_GRSL2022} comparing $135$ different Haralick-based descriptors to define which bands should to be adopted to represent those segments.

\subsection{Learning Methods}
In order to compare the performance of the e-NEAT approach, twelve different machine learning methods were applied, covering both single classifiers and ensemble methods. Regarding the single classifiers, were compared MLP Classifier~\cite{mlp1994}, K-Nearest Neighbors~\cite{Friedman2001}, Gaussian NB~\cite{nbt_1996}, Decision Tree~\cite{dt-survey}, and three SVM-based classifiers using linear, polynomial, and RBF kernels~\cite{svm}. For the ensemble methods, were employed: AdaBoost~\cite{adaboost}, Random Forest\cite{randomforest}, Bagging Classifier\cite{bagging}, Extra Trees and Gradient Boosting\cite{friedman2001greedy}. All of learning methods were implemented through the \textit{Scikit-learn} \cite{skLearn} library, without any changes in their default parameters.

\subsection{Evaluation Measure}
Since the database used in this paper is highly unbalanced between the Forest and Non-forest classes, the balanced accuracy measure ($B_{Acc}$) has been adopted to compute the real performance of the methods. The balanced accuracy equation for binary classification problems is defined by: 
\begin{equation}
B_{Acc} = \frac{1}{2} \left ( \frac{TP}{TP + FN} + \frac{TN}{TN + FP} \right )
\textit{ ,}
\label{eq:balancedAccuracy}
\end{equation}
where, $TP$ is equal to the number true positives; $TN$ indicates the number of true negatives; $FP$ corresponds to false positives, and $FN$ is equal to the number of false negatives. The score ranges from 0 to 1, the higher the score, the greater the performance.

\subsection{Experimental Setup}

The e-NEAT method uses a NEAT implementation called \textit{neat-python}~\cite{neatPython} to perform experiments with the neuroevolution technique. The main set of parameters defined in the NEAT technique are: 
(1) population size is equal to $200$; (2) number of generation is equal to $75$; (3) fitness value is equal to $100\%$ of balanced accuracy; (4) activation functions (sigmoid, tanh, relu, log, clamped, hat, identity, and softplus); (5) neurons in the hidden layer is equal to $8$; and (6) network configuration is fully conected. Additionally, in the proposed e-NEAT method one more parameter needs to define: number of ANNs that compose the ensemble is equal to $15$. All parameters were found through an exhaustive set of experiments carried out, which take into account the trade-off between effectiveness (maximize balanced accuracy) and efficiency (minimize time consumption).

\section{Result and Discussion}
In this section, three comparative analyses were performed in a five-round protocol and the balanced accuracy measure was computed for each of them. First, an analysis among single classifiers baselines and the best ANN-based NEAT technique. Second, a comparison among the ensemble baseline methods and the proposed e-NEAT. Third, an analysis of the performance of ANNs based on NEAT technique. %Finally, a fine-grained analysis among the best ensemble method. 

\subsection{Analysis of Single Classifiers}

Figure~\ref{fig:NonEnsembleMethodsPerformance} shows the classification results of seven single baseline classifiers existing in the literature and the best ANN based on NEAT technique. As can be observed, the best ANN classifier created by NEAT technique achieved $91.5\%$ of balanced accuracy against $88.7\%$ achieved by the best baseline method (SVM using linear kernel) proposed in~\cite{Dallaqua_GRSL2022}. 

\begin{figure}[!h]
\begin{center}
\includegraphics[width=0.48\textwidth]{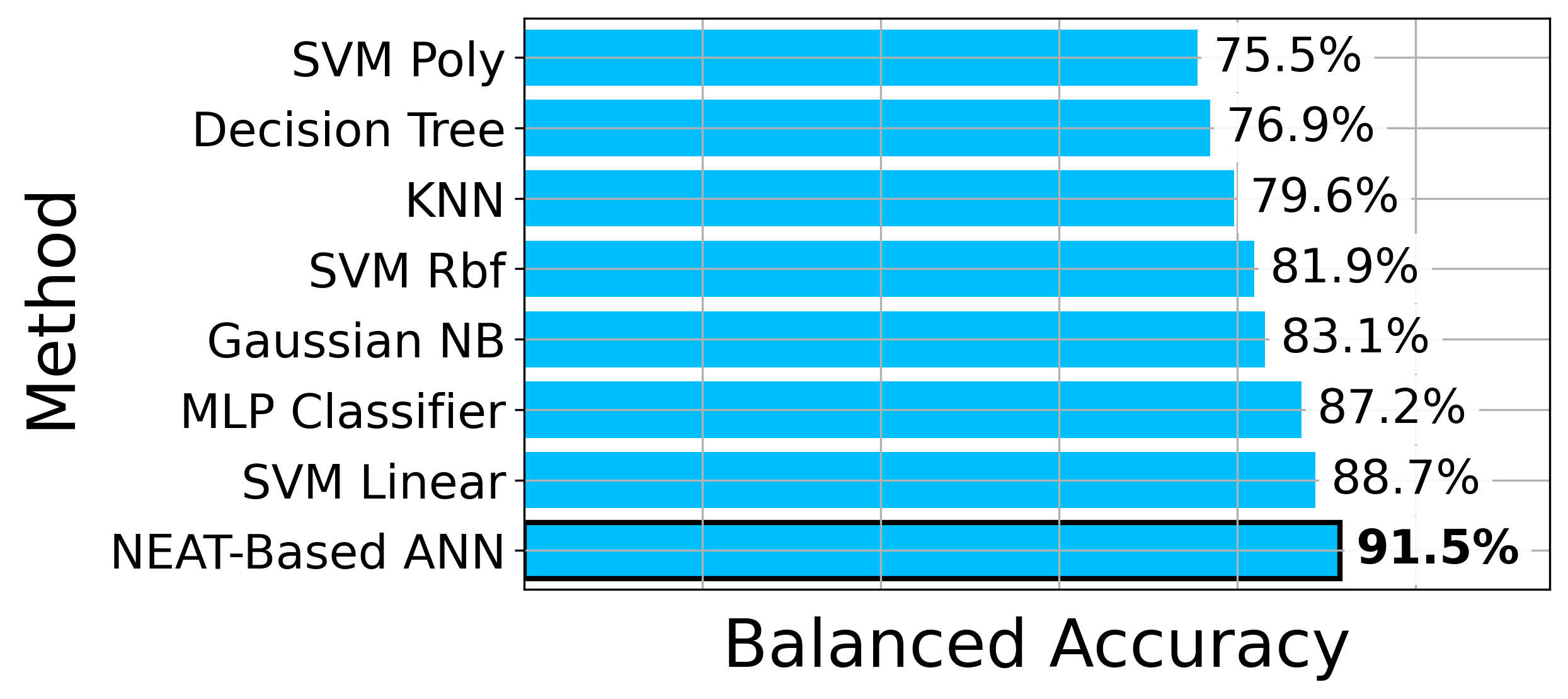}
\end{center}
\caption{The best classification results for all of baseline single classifiers compared with the ANN based on NEAT technique.}
\label{fig:NonEnsembleMethodsPerformance}
\end{figure}

\subsection{Analysis of Ensemble Methods}

Table~\ref{tab:ensemble} shows the classification results of four different baseline methods existing in literature and the e-NEAT method proposed in this paper. As can be observed, e-NEAT method achieved the best results against all of baseline methods. Furthermore, e-NEAT method showed to be the best ensemble of classifier with $89.6\%$ of balanced accuracy against $84.4\%$ achieved by the best baseline method (Bagging Classifier) using the same amount of base classifiers ($15$), representing $6.2\%$ of relative gain between them. It is very important to note that e-NEAT even reduces standard deviation (Std) value, as well as, it achieves greater balanced accuracies in minimal (Min.) and maximal (Max.) values.

\begin{table}[ht!]
\centering
\caption{Classification results of five different ensemble methods baselines and our proposed ensemble methods (e-NEAT) for $5$ rounds of performed experiments.}
\resizebox{8.4cm}{!}{
\begin{tabular}{lrrrr}
\toprule
\multirow{2}{*}{\textbf{Ensemble Methods}} &  \multicolumn{4}{c}{\textbf{Balanced Accuracy}} \\
\cline{2-5}
 &   \textbf{Mean} &  \textbf{Std} &    \textbf{Min.} &    \textbf{Max.} \\
\toprule
% new experiment:
AdaBoost                &    47.2 &  0.0 &  47.2 &  47.2    \\
%Bagging \textbf{(best baseline)}   &    83.3 &  4.6 &  75.8 &  86.7    \\
Bagging \textbf{(best baseline)}   &    84.4 &  3.7 &  78.8 &  89.7    \\
Extra Trees             &    81.9 &  4.2 &  76.0 &  87.1    \\
Gradient Boosting       &    79.4 &  1.2 &  78.3 &  81.1    \\
Random Forest           &    81.5 &  3.9 &  77.9 &  87.9    \\ 
\toprule
\textbf{e-NEAT}    &   \textbf{89.6} &  \textbf{0.9} &  \textbf{88.3} &  \textbf{90.7}    \\
\bottomrule
\end{tabular}
}
\label{tab:ensemble}
\end{table}

In literature, a method can be considered a good ensemble of classifiers  whether it performance at least above the average of its base classifiers in a target task. Therefore, Figure~\ref{fig:ANNsAndEnsemblePerformance} presents the classification results extracted from one of the five rounds of the performed experiments, showing the performance of fifteen ANNs individually, as well as, the aggregation strategy (e-NEAT) which combines them. Notice that the pool of ANNs achieved an average balanced accuracy of $81.4\%$, while the e-NEAT method obtained $90.7\%$. Therefore, it means that, on average, the classification results achieved by ANNs (individually created by NEAT) are $11.4\%$ (relative gain) worse than the ensemble method proposed in this work (e-NEAT).

\begin{figure}[!h]
\begin{center}
\includegraphics[width=0.4\textwidth]{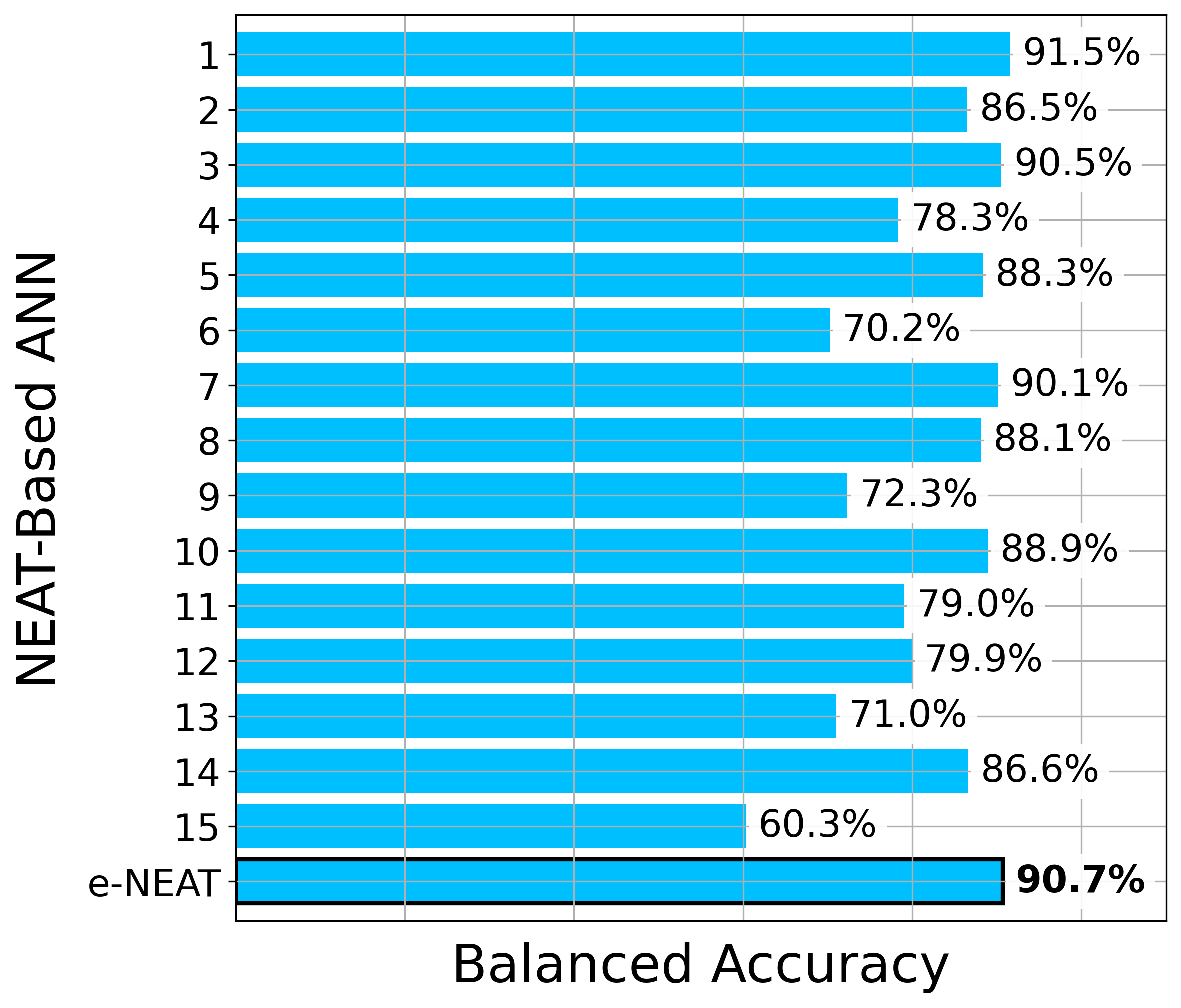}
\end{center}
\caption{Classification results of the fifteen different ANNs created by NEAT technique and their combination method represented by e-NEAT for only one round of the experiments. The mean of balanced accuracy and standard deviation of ANN-based NEAT are $81.4\%$ and $9.4\%$, respectively. }
\label{fig:ANNsAndEnsemblePerformance}
\end{figure}

\section{Conclusion}
\label{s.conclusion}
In this paper, a framework for combining artificial neural networks based on neuroevolution technique (NEAT) has been proposed. Furthermore, the aggregation strategy has been developed to combine the NEAT-based ANNs and to improve the classification results in the recent deforestation detection task. In performed experiments it was possible to see that the NEAT technique is able to create more diverse ANNs to be combined by the aggregation strategy. In addition, the e-NEAT approach, which uses the mode as aggregation strategy, achieved the best results with a relative gain of $7.6\%$ against the best ensemble baseline compared in this paper. Finally, it was possible to observe that the proposed e-NEAT method might be a good solution for classifying image segments in scenarios of reduced training set as the recent deforestation detection task.   

%\begin{figure}[!ht]
%\begin{tabular}{c}
%\includegraphics[width=0.27\textwidth]{results/confusionMatrix_e-NEAT_mode.png} \\ (a) Forest = 93.0\%, Non-forest = 88.9\% and $B_{Acc} = 91.0\%$.  \\
%\includegraphics[width=0.28\textwidth]{results/confusionMatrix_BaggingClassifier.png} \\
%    (b) Forest = 93.9\%, Non-forest = 81.7\% and $B_{Acc} = 87.8\%$.  
%\end{tabular}
%\caption{In (a) is the classification results of the e-NEAT method. In (b) is the results of Bagging Classifier method.}
%\label{fig:confusionMatrixENeat}

%\end{figure}

\section*{Acknowledgements}
Authors would like to thank the research funding agencies CAPES, CNPq through the Universal Project (grant \#408919/2016-7). This research is part of the INCT of the Future Internet for Smart Cities funded by CNPq (grant \#465446/2014-0), CAPES (Finance Code 001), and FAPESP (grants \#2014/50937-1, \#2015/24485-9, \#2017/25908-6, and \#2018/23908-1 ).

\bibliographystyle{IEEEtran}
\bibliography{refs}

\end{document}